\documentclass{article}

\usepackage{arxiv}

\usepackage[utf8]{inputenc} 
\usepackage[T1]{fontenc}    
\usepackage{hyperref}       
\usepackage{url}            
\usepackage{booktabs}       
\usepackage{amsfonts}       
\usepackage{nicefrac}       
\usepackage{microtype}      
\usepackage{lipsum}		
\usepackage{graphicx}
\usepackage{natbib}
\usepackage{doi}
\usepackage{wrapfig}
\usepackage{color}

\title{RT-DETRv2: Improved Baseline with Bag-of-Freebies for Real-Time Detection Transformer}

\author{
\vspace{1mm}
Wenyu Lv\textsuperscript{1} \quad Yian Zhao\textsuperscript{2} \quad
Qinyao Chang\textsuperscript{1} \quad Kui Huang\textsuperscript{1} \quad Guanzhong Wang\textsuperscript{1} \quad Yi Liu\textsuperscript{1} \\
\vspace{1mm}
\small\textsuperscript{1}Baidu Inc. \quad 
\small\textsuperscript{2}Peking University Shenzhen Graduate School\\
\small \href{mailto:lvwenyu01@baidu.com}{lvwenyu01@baidu.com} \quad
\small \href{mailto:zhaoyian@stu.pku.edu.cn}{zhaoyian@stu.pku.edu.cn}  
}

\renewcommand{\undertitle}{Technical Report}
\renewcommand{\shorttitle}{RT-DETRv2: Improved Baseline with Bag-of-Freebies for Real-Time Detection Transformer}
\definecolor{brickred}{rgb}{0.8, 0.25, 0.33}
\definecolor{ao(english)}{rgb}{0.0, 0.5, 0.0}
\newcommand{\increase}[1]{{
  \fontsize{8.5pt}{0.5em}\selectfont({\color{ao(english)}{$\uparrow$~\textbf{#1}}})
}}
\newcommand{\decrease}[1]{{
  \fontsize{7.5pt}{0.5em}\selectfont({\color{brickred}{$\downarrow$~\textbf{#1}}})
}}

\begin{document}
\maketitle

\begin{abstract}
In this report, we present RT-DETRv2, an improved Real-Time DEtection TRansformer~(RT-DETR).
RT-DETRv2 builds upon the previous state-of-the-art real-time detector, RT-DETR, and opens up a set of bag-of-freebies for flexibility and practicality, as well as optimizing the training strategy to achieve enhanced performance.
To improve the flexibility, we suggest setting a distinct number of sampling points for features at different scales in the deformable attention to achieve selective multi-scale feature extraction by the decoder.
To enhance practicality, we propose an optional discrete sampling operator to replace the \texttt{grid\_sample} operator that is specific to RT-DETR compared to YOLOs. This removes the deployment constraints typically associated with DETRs.
For the training strategy, we propose dynamic data augmentation and scale-adaptive hyperparameters customization to improve performance without loss of speed.
Source code and pre-trained models will be available at \href{https://github.com/lyuwenyu/RT-DETR}{https://github.com/lyuwenyu/RT-DETR}.
\end{abstract}

\section{Introduction}
Object detection is a fundamental vision task that involves identifying and localizing objects in an image.
Among them, real-time object detection is an important field and has a wide range of applications, such as autonomous driving~(\cite{atakishiyev2024explainable}).
With the development of the last few years, YOLO detectors~(\cite{redmon2017yolo9000,redmon2018yolov3,bochkovskiy2020yolov4,yolov5v7.0,xu2022pp,li2023yolov6v3,wang2023yolov7,yolov8,wang2024yolov9,wang2024yolov10}) are without doubt the most prestigious framework in this field.
The reason for this is the reasonable balance achieved by the YOLO detectors.

The advent of RT-DETR~(\cite{zhao2024detrs}) opens up a new technological avenue for real-time object detection, breaking the dependency on the YOLO in this field.
RT-DETR proposes an efficient hybrid encoder to replace the vanilla Transformer encoder in DETR~(\cite{carion2020end}), which significantly improves the inference speed by decoupling the intra-scale interaction and cross-scale fusion of multi-scale features.
To further improve the performance, RT-DETR proposes the uncertainty-minimal query selection, which provides high-quality initial queries to the decoder by explicitly optimizing the uncertainty.
Moreover, RT-DETR provides a wide range of detector sizes and supports flexible speed tuning to accommodate various real-time scenarios without retraining.
RT-DETR represents a novel, end-to-end, real-time detector that marks a significant advancement for the DETR family.

In this report, we present RT-DETRv2, an improved real-time detection Transformer.
This work is built upon the recent RT-DETR and opens up a set of bag-of-freebies for flexibility and practicality within the DETR family, as well as optimizing the training strategy to achieve enhanced performance.
Specifically, RT-DETRv2 suggests setting a distinct number of sampling points for features at different scales within the deformable attention module to achieve selective multi-scale feature extraction by the decoder.
In the realm of enhancing practicality, RT-DETRv2 provides an optional discrete sampling operator to replace the original \texttt{grid\_sample} operator, which is specific to DETRs, thus eliminating the deployment constraints typically associated with detection Transformers.
Furthermore, RT-DETRv2 optimizes the training strategy, including dynamic data augmentation and scale-adaptive hyperparameters customization, with the objective of improving performance without loss of speed.
The results demonstrate that RT-DETRv2 provides an improved baseline with bag-of-freebies for RT-DETR, increases the flexibility and practicality, and the proposed training strategies optimize the performance and training cost.

\section{Method}
The framework of RT-DETRv2 remains the same as RT-DETR, with only modifications to the deformable attention module of the decoder.

\subsection{Framework}
\noindent \textbf{Distinct number of sampling points for different scales.} 
Current DETRs utilize the deformable attention module~(\cite{zhu2020deformable}) to alleviate the high computational overhead caused by the long sequence of inputs composed of multi-scale features.
The RT-DETR decoder retains this module, which defines the same number of sampling points at each scale.
We argue that this constraint ignores the intrinsic differences in features at different scales and limits the feature extraction capability of the deformable attention module.
Therefore, we propose to set distinct numbers of sampling points for different scales to achieve more flexible and efficient feature extraction.

\noindent \textbf{Discrete sampling.}
To improve the practicality of the RT-DETR and to make it available everywhere.
We focus on comparing the deployment requirements of YOLOs and RT-DETR, where the RT-DETR-specific \texttt{grid\_sample} operator limits its broad applicability.
Therefore, we propose an optional \texttt{discrete\_sample} operator to replace the \texttt{grid\_sample}, thus removing the deployment constraints of RT-DETR.
Specifically, we perform a rounding operation on the predicted sampling offsets, omitting the time-consuming bilinear interpolation. 
However, the rounding operation is non-differentiable, so we turn off the gradient of the parameters used to predict the sampling offsets.
In practice, we first employ the \texttt{grid\_sample} operator for training and then replace it with the \texttt{discrete\_sample} operator for fine-tuning.
For inference and deployment, the model employs the \texttt{discrete\_sample} operator.

\subsection{Training Scheme}
\noindent \textbf{Dynamic data augmentation.} 
To equip the model with robust detection performance, we propose the dynamic data augmentation strategy. 
Considering the poor generalizability of the detector in the early training period, we apply stronger data augmentation, while in the later training period we decrease its level to adapt the detector to the detection of the target domain.
Specifically, we maintain the RT-DETR data augmentation in the early period, while turning off \texttt{RandomPhotometricDistort}, \texttt{RandomZoomOut}, \texttt{RandomIoUCrop}, and \texttt{MultiScaleInput} in the last two epochs.

\noindent \textbf{Scale-adaptive hyperparameters customization.}
We also observe that the scaled RT-DETRs of different sizes are trained with the same optimizer hyperparameters, resulting in their sub-optimal performance.
Therefore, we propose scale-adaptive hyperparameters customization for scaled RT-DETRs.
Considering that the pre-trained backbone for light detector (\textit{e.g.}, ResNet18~(\cite{he2016deep})) has lower feature quality, we increase its learning rate. On the contrary, the pre-trained backbone with large detector (\textit{e.g.}, ResNet101~(\cite{he2016deep})) has higher feature quality and we decrease its learning rate.


\section{Experiment}
\begin{wraptable}[9]{r}{7.0cm}
\vspace{-6.5mm}
\caption{The hyperparameters of RT-DETRv2.}
\vspace{1mm}
\label{tab:hyper}
\centering
\renewcommand{\arraystretch}{1.15}
\setlength{\tabcolsep}{5.33pt}
\begin{tabular}{cccc}
  \toprule
   \textbf{Model} & \textbf{Backbone} & \textit{lr$_{backbone}$} & \textit{lr$_{det}$} \\ 
  \midrule
  \midrule
  RT-DETRv2-S & ResNet18 & 1e-4 & 1e-4 \\
  RT-DETRv2-M & ResNet34 & 5e-5 & 1e-4 \\
  RT-DETRv2-L & ResNet50 & 1e-5 & 1e-4 \\
  RT-DETRv2-X & ResNet101 & 1e-6 & 1e-4 \\
  \bottomrule
\end{tabular}
\end{wraptable}
\subsection{Implementation Details}
As with RT-DETR, we use ResNet~(\cite{he2016deep}) pretrained on ImageNet as the backbone and train RT-DETRv2 with the AdamW~(\cite{loshchilov2018decoupled}) optimizer with a batch size of $16$ and apply the exponential moving average~(EMA) with $ema\_decay=0.9999$.
For the optional discrete sampling, we first pre-train $6\times$ with the \texttt{grid\_sample} operator and then fine-tune $1\times$ with the \texttt{discrete\_sample} operator.
For scale-adaptive hyperparameters customization, the hyperparameters are shown in Tab.~\ref{tab:hyper}, where \textit{lr} represents the learning rate.

\subsection{Evaluation}
RT-DETRv2 is trained on COCO~(\cite{lin2014microsoft}) \texttt{train2017} and validated on COCO \texttt{val2017} dataset.
We report the standard AP metrics~(averaged over uniformly sampled IoU thresholds ranging from $0.50-0.95$ with a step size of $0.05$), and AP$^{val}_{50}$ commonly used in real scenarios.

\subsection{Results}
The comparison with RT-DETR(\cite{zhao2024detrs}) is shown in Tab.~\ref{tab:table}.
RT-DETRv2 outperforms RT-DETR at different scales of detectors without loss of speed.
\begin{table}[ht]
\caption{\textbf{Comparison of RT-DETR and RT-DETRv2.} The FPS is reported on T4 GPU with TensorRT FP16. For evaluation, all input sizes are fixed on 640 $\times$ 640.}
\centering
\begin{tabular}{llccccc}
\toprule
\textbf{Model}  & \textbf{Backbone}  & \textbf{Dataset} & \textbf{\#Params~(M)} & \textbf{FPS$_{bs=1}$} & \textbf{AP$^{val}$} & \textbf{AP$^{val}_{50}$} \\
\midrule
RT-DETR-S & ResNet18 & COCO & 20 & 217 & 46.5 & 63.8 \\
RT-DETR-M & ResNet34 & COCO & 31 & 161 & 48.9 & 66.8 \\
RT-DETR-M$^*$ & ResNet50 & COCO & 36 & 145 & 51.3 & 69.6 \\
RT-DETR-L & ResNet50 & COCO & 42 & 108 & 53.1 & 71.3 \\
RT-DETR-X & ResNet101 & COCO & 76 & 74 & 54.3 & 72.7 \\
\midrule
RT-DETRv2-S & ResNet18 & COCO & 20 & 217 & 47.9 \increase{1.4} & 64.9 \increase{1.1} \\
RT-DETRv2-M & ResNet34 & COCO & 31 & 161 & 49.9 \increase{1.0} & 67.5 \increase{0.7} \\
RT-DETRv2-M$^*$ & ResNet50 & COCO & 36 & 145 & 51.9 \increase{0.6} & 69.9 \increase{0.3} \\
RT-DETRv2-L & ResNet50 & COCO & 42 & 108 & 53.4 \increase{0.3} & 71.6 \increase{0.3} \\
RT-DETRv2-X & ResNet101 & COCO & 76 & 74 & 54.3 \increase{0.0} & 72.8 \increase{0.1} \\
\bottomrule
\end{tabular}
\label{tab:table}
\end{table}

\subsection{Ablations}
\noindent \textbf{Ablation on sampling points.}
We perform an ablation study on the total number of sampling points of the \texttt{grid\_sample} operator.
The total number of sampling points is calculated as $\texttt{num\_head} \times \texttt{num\_point} \times \texttt{num\_query} \times \texttt{num\_decoder}$, where $\texttt{num\_point}$ represents the sum of sampling points for each scale feature in each grid.
The results show that reducing the number of sampling points does not cause a significant degradation in the performance, \textit{cf.}~Tab.~\ref{tab:point}.
This means that practical application is unlikely to be affected in most industrial scenarios.

\begin{table}[ht]
\caption{{Ablation on sampling points.}}
\centering
\begin{tabular}{lcccc}
\toprule
\textbf{Model} & \textbf{Sampling method} & \textbf{\#Points} & \textbf{AP$^{val}$} & \textbf{AP$^{val}_{50}$} \\
\midrule
RT-DETRv2-S & \texttt{grid\_sample} & 86,400 & 47.9 & 64.9 \\
RT-DETRv2-S & \texttt{grid\_sample} & 64,800 & 47.8 & 64.8 \decrease{0.1}\\
RT-DETRv2-S & \texttt{grid\_sample} & 43,200 & 47.7 & 64.7 \decrease{0.2}\\
RT-DETRv2-S & \texttt{grid\_sample} & 21,600 & 47.3 & 64.3 \decrease{0.6}\\
\bottomrule
\end{tabular}
\label{tab:point}
\end{table}

\noindent \textbf{Ablation on discrete sampling.}
We then remove the \texttt{grid\_sample} and replace it with \texttt{discrete\_sample} for the ablation. The results show that this operation does not cause a noticeable reduction in AP$^{val}_{50}$, but does eliminate the deployment constraints of the DETRs, \textit{cf.}~Tab.~\ref{tab:discrete}.

\begin{table}[ht]
\caption{{Ablation on discrete sampling.}}
\centering
\begin{tabular}{lcccc}
\toprule
\textbf{Model} & \textbf{Backbone} & \textbf{Sampling method} & \textbf{AP$^{val}$} & \textbf{AP$^{val}_{50}$} \\
\midrule
RT-DETRv2-S & ResNet18 & \texttt{discrete\_sample} & 47.4 & 64.8 \decrease{0.1} \\
RT-DETRv2-M & ResNet34 & \texttt{discrete\_sample} & 49.2 & 67.1 \decrease{0.4} \\
RT-DETRv2-M$^*$ & ResNet50 & \texttt{discrete\_sample} & 51.4 & 69.7 \decrease{0.2} \\
RT-DETRv2-L & ResNet50 & \texttt{discrete\_sample} & 52.9 & 71.3 \decrease{0.3} \\
\bottomrule
\end{tabular}
\label{tab:discrete}
\end{table}

\section{Conclusion}
In this report, we propose RT-DETRv2, an improved real-time detection Transformer.
RT-DETRv2 opens up a set of bag-of-freebies to increase the flexibility and practicality of RT-DETR, optimizing the training strategy to achieve enhanced performance without loss of speed.
We hope that this report will provide insights for the DETR family and broaden the scope of RT-DETR applications.

{
    \bibliographystyle{unsrtnat}
    \bibliography{main}

\begin{thebibliography}{17}
\providecommand{\natexlab}[1]{#1}
\providecommand{\url}[1]{\texttt{#1}}
\expandafter\ifx\csname urlstyle\endcsname\relax
  \providecommand{\doi}[1]{doi: #1}\else
  \providecommand{\doi}{doi: \begingroup \urlstyle{rm}\Url}\fi

\bibitem[Atakishiyev et~al.(2024)Atakishiyev, Salameh, Yao, and Goebel]{atakishiyev2024explainable}
Shahin Atakishiyev, Mohammad Salameh, Hengshuai Yao, and Randy Goebel.
\newblock Explainable artificial intelligence for autonomous driving: A comprehensive overview and field guide for future research directions.
\newblock \emph{IEEE Access}, 2024.

\bibitem[Redmon and Farhadi(2017)]{redmon2017yolo9000}
Joseph Redmon and Ali Farhadi.
\newblock Yolo9000: better, faster, stronger.
\newblock In \emph{Proceedings of the IEEE/CVF Conference on Computer Vision and Pattern Recognition}, pages 7263--7271, 2017.

\bibitem[Redmon and Farhadi(2018)]{redmon2018yolov3}
Joseph Redmon and Ali Farhadi.
\newblock Yolov3: An incremental improvement.
\newblock \emph{arXiv preprint arXiv:1804.02767}, 2018.

\bibitem[Bochkovskiy et~al.(2020)Bochkovskiy, Wang, and Liao]{bochkovskiy2020yolov4}
Alexey Bochkovskiy, Chien-Yao Wang, and Hong-Yuan~Mark Liao.
\newblock Yolov4: Optimal speed and accuracy of object detection.
\newblock \emph{arXiv preprint arXiv:2004.10934}, 2020.

\bibitem[Glenn.(2022)]{yolov5v7.0}
Jocher Glenn.
\newblock Yolov5 release v7.0.
\newblock \emph{\url{https://github.com/ultralytics/yolov5/tree/v7.0}}, 2022.

\bibitem[Xu et~al.(2022)Xu, Wang, Lv, Chang, Cui, Deng, Wang, Dang, Wei, Du, et~al.]{xu2022pp}
Shangliang Xu, Xinxin Wang, Wenyu Lv, Qinyao Chang, Cheng Cui, Kaipeng Deng, Guanzhong Wang, Qingqing Dang, Shengyu Wei, Yuning Du, et~al.
\newblock Pp-yoloe: An evolved version of yolo.
\newblock \emph{arXiv preprint arXiv:2203.16250}, 2022.

\bibitem[Li et~al.(2023)Li, Li, Geng, Jiang, Cheng, Zhang, Ke, Xu, and Chu]{li2023yolov6v3}
Chuyi Li, Lulu Li, Yifei Geng, Hongliang Jiang, Meng Cheng, Bo~Zhang, Zaidan Ke, Xiaoming Xu, and Xiangxiang Chu.
\newblock Yolov6 v3.0: A full-scale reloading.
\newblock \emph{arXiv preprint arXiv:2301.05586}, 2023.

\bibitem[Wang et~al.(2023)Wang, Bochkovskiy, and Liao]{wang2023yolov7}
Chien-Yao Wang, Alexey Bochkovskiy, and Hong-Yuan~Mark Liao.
\newblock Yolov7: Trainable bag-of-freebies sets new state-of-the-art for real-time object detectors.
\newblock In \emph{Proceedings of the IEEE/CVF Conference on Computer Vision and Pattern Recognition}, pages 7464--7475, 2023.

\bibitem[Glenn.(2023)]{yolov8}
Jocher Glenn.
\newblock Yolov8.
\newblock \emph{\url{https://github.com/ultralytics/ultralytics/tree/main}}, 2023.

\bibitem[Wang et~al.(2024{\natexlab{a}})Wang, Yeh, and Liao]{wang2024yolov9}
Chien-Yao Wang, I-Hau Yeh, and Hong-Yuan~Mark Liao.
\newblock Yolov9: Learning what you want to learn using programmable gradient information.
\newblock \emph{arXiv preprint arXiv:2402.13616}, 2024{\natexlab{a}}.

\bibitem[Wang et~al.(2024{\natexlab{b}})Wang, Chen, Liu, Chen, Lin, Han, and Ding]{wang2024yolov10}
Ao~Wang, Hui Chen, Lihao Liu, Kai Chen, Zijia Lin, Jungong Han, and Guiguang Ding.
\newblock Yolov10: Real-time end-to-end object detection.
\newblock \emph{arXiv preprint arXiv:2405.14458}, 2024{\natexlab{b}}.

\bibitem[Zhao et~al.(2024)Zhao, Lv, Xu, Wei, Wang, Dang, Liu, and Chen]{zhao2024detrs}
Yian Zhao, Wenyu Lv, Shangliang Xu, Jinman Wei, Guanzhong Wang, Qingqing Dang, Yi~Liu, and Jie Chen.
\newblock Detrs beat yolos on real-time object detection.
\newblock In \emph{Proceedings of the IEEE/CVF Conference on Computer Vision and Pattern Recognition}, pages 16965--16974, 2024.

\bibitem[Carion et~al.(2020)Carion, Massa, Synnaeve, Usunier, Kirillov, and Zagoruyko]{carion2020end}
Nicolas Carion, Francisco Massa, Gabriel Synnaeve, Nicolas Usunier, Alexander Kirillov, and Sergey Zagoruyko.
\newblock End-to-end object detection with transformers.
\newblock In \emph{European Conference on Computer Vision}, pages 213--229. Springer, 2020.

\bibitem[Zhu et~al.(2020)Zhu, Su, Lu, Li, Wang, and Dai]{zhu2020deformable}
Xizhou Zhu, Weijie Su, Lewei Lu, Bin Li, Xiaogang Wang, and Jifeng Dai.
\newblock Deformable detr: Deformable transformers for end-to-end object detection.
\newblock In \emph{International Conference on Learning Representations}, 2020.

\bibitem[He et~al.(2016)He, Zhang, Ren, and Sun]{he2016deep}
Kaiming He, Xiangyu Zhang, Shaoqing Ren, and Jian Sun.
\newblock Deep residual learning for image recognition.
\newblock In \emph{Proceedings of the IEEE conference on computer vision and pattern recognition}, pages 770--778, 2016.

\bibitem[Loshchilov and Hutter(2018)]{loshchilov2018decoupled}
Ilya Loshchilov and Frank Hutter.
\newblock Decoupled weight decay regularization.
\newblock In \emph{International Conference on Learning Representations}, 2018.

\bibitem[Lin et~al.(2014)Lin, Maire, Belongie, Hays, Perona, Ramanan, Doll{\'a}r, and Zitnick]{lin2014microsoft}
Tsung-Yi Lin, Michael Maire, Serge Belongie, James Hays, Pietro Perona, Deva Ramanan, Piotr Doll{\'a}r, and C~Lawrence Zitnick.
\newblock Microsoft coco: Common objects in context.
\newblock In \emph{European Conference on Computer Vision}, pages 740--755. Springer, 2014.

\end{thebibliography}
}

\end{document}